\documentclass[pmlr,twocolumn,10pt]{jmlr} 

\RestyleAlgo{ruled}
\SetAlgoVlined
\LinesNumbered

\usepackage{xcolor}
\usepackage{colortbl}
\usepackage{booktabs}
\usepackage{multirow}
\usepackage{adjustbox}
\usepackage{array}
\usepackage{graphicx}
\usepackage{rotating}
\usepackage{float}
\usepackage{amssymb}

\newcommand{\PreserveBackslash}[1]{\let\temp=\\#1\let\\=\temp}
\newcolumntype{C}[1]{>{\PreserveBackslash\centering}p{#1}}
\newcolumntype{R}[1]{>{\PreserveBackslash\raggedleft}p{#1}}
\newcolumntype{L}[1]{>{\PreserveBackslash\raggedright}p{#1}}
\newcolumntype{P}[1]{>{\raggedright\arraybackslash}m{#1}}

\usepackage{tikz}
\usetikzlibrary{shapes, arrows.meta, positioning, fit, backgrounds, calc}

\usepackage{siunitx}

\usepackage[switch]{lineno}

\theorembodyfont{\upshape}
\theoremheaderfont{\scshape}
\theorempostheader{:}
\theoremsep{\newline}

\jmlrvolume{333}
\jmlryear{2026}
\jmlrsubmitted{}
\jmlrpublished{}
\jmlrworkshop{Conference on Health, Inference, and Learning (CHIL) 2026} 

\title[Survey-aware Machine Learning]{Survey-aware Machine Learning: A Guideline for Valid Population Health Inference based on Scoping Review}

\author{%
\Name{YongKyung Oh}
\Email{yongkyungoh@mednet.ucla.edu}\\
\addr Medical \& Imaging Informatics (MII), UCLA
\AND
\Name{Henry W. Zheng}
\Email{henryzheng@ucla.edu}\\
\addr Medical \& Imaging Informatics (MII), UCLA
\AND
\Name{Jeffrey Feng}
\Email{j64feng@ucla.edu}\\
\addr Medical \& Imaging Informatics (MII), UCLA
\AND
\Name{Alex A. T. Bui}\thanks{Corresponding Author}
\Email{buia@mii.ucla.edu}\\
\addr Medical \& Imaging Informatics (MII), UCLA
}

\begin{document}

\maketitle

\begin{abstract}
    Machine Learning (ML) models trained on complex health surveys such as the National Health and Nutrition Examination Survey (NHANES) often ignore primary sampling units, stratification variables, and sampling weights. This practice violates the independence assumptions of standard evaluation methods. As a result, estimates become biased, uncertainty is underestimated, and fairness assessments fail to reflect population-level disparities. We propose Survey-aware Machine Learning (SaML), a nine-step guideline that incorporates survey design metadata across the ML lifecycle. Through a scoping review of 16 methodological papers, we summarize existing work on weighted model training, design-based cross-validation, and survey-adjusted performance evaluation. We also identify gaps in hyperparameter tuning and deployment. We provide task-specific guidance that clarifies which steps are required for different analytical objectives. SaML provides a checklist for valid population inference from survey data.
\end{abstract}

\paragraph*{Data and Code Availability}
This paper presents a systematic audit and methodological review of peer-reviewed literature. The empirical illustration (Appendix~\ref{sec:appendix_empirical}) uses the National Health and Nutrition Examination Survey (NHANES) 2021--2023 cycle, which is publicly available from the CDC.\footnote{\url{https://www.cdc.gov/nchs/nhanes/}} R code for data preprocessing and all experiments is available at \url{https://github.com/yongkyung-oh/SaML}. 

All analyzed studies and synthesized metadata are cited in the references and are accessible through their respective journals and data repositories.

\paragraph*{Author Contributions}
Y.O. conceived the study, designed the SaML framework, conducted the review and analysis, and drafted the manuscript. 
H.Z. and J.F. contributed to screening, framework refinement, result interpretation, and revision. 
A.B. supervised the study, contributed to study design and interpretation, and revised the manuscript. 

\paragraph*{Institutional Review Board (IRB)}
The research presented in this paper consists of a systematic audit of published methodological practices and synthesis of existing literature. It does not involve interaction with human subjects or the use of private, identifiable individual data. 
According to U.S. Department of Health and Human Services guidelines, this study is categorized as Not Human Subject Research and did not require IRB approval.

\section{Introduction}

Recent advances in AI-assisted research systems seek to accelerate scientific discovery in biomedicine and population health \citep{krenn_scientific_2022, wang_scientific_2023, taha_machine_2025}. The reliability of these systems depends on the validity of evaluation methods used to assess model performance on observational data \citep{kapoor_leakage_2023, lones_avoiding_2024}. 
As machine learning (ML) models inform clinical decisions and health policy, they encounter data that violate the independent and identically distributed (i.i.d.) assumption through stratification, clustering, and unequal selection probabilities \citep{pfeffermann_role_1993,lohr_sampling_2021}. 
Standard ML evaluation pipelines often remain survey-naive and treat complex samples as simple random draws.
As a result, sample-level and population-level estimands diverge, but this distinction is rarely stated explicitly.

This mismatch is pronounced in analyses of complex surveys such as the National Health and Nutrition Examination Survey (NHANES) and the Behavioral Risk Factor Surveillance System (BRFSS). These datasets rely on stratification to ensure subgroup representation, clustering to reduce cost, and unequal weights to adjust for selection probabilities \citep{lohr_sampling_2021}. Ignoring survey design leads to biased parameter estimates \citep{pfeffermann_role_1993}, underestimated variance \citep{binder_variances_1983}, and misleading fairness assessments \citep{schuch_fairness_2023}.

To address these issues, we propose \textbf{Survey-aware Machine Learning (SaML)} as a methodological guardrail for AI-assisted health research. This paper makes three contributions.
\begin{enumerate}
    \item \textbf{Guideline for population-valid ML}. We present a nine-step guideline that ensures ML models trained on complex health surveys such as NHANES and BRFSS produce population-level estimates rather than sample-level results.

    \item \textbf{Evidence synthesis}. We conduct a scoping review of 16 methodological papers and document fragmented progress. Existing work addresses individual steps, but no complete pipeline exists. Hyperparameter tuning and deployment remain unresolved gaps for building trustworthy models.

    \item \textbf{Actionable task mapping}. We introduce a framework that identifies which survey-aware steps are required for a given ML task. This mapping reduces unnecessary complexity while preserving methodological validity.
\end{enumerate}
The remainder of this paper is organized as follows. Section~\ref{sec:motivating_ex} illustrates the impact of survey-aware methods using NHANES data. Section~\ref{sec:related_work} reviews prior work on survey methodology and health ML. Section~\ref{sec:guide} presents the nine-step SaML guideline. Section~\ref{sec:review} reports a scoping review that identifies coverage and gaps across the nine steps and provides task-specific guidance. Section~\ref{sec:implication} discusses implications for reproducibility, fairness, and implementation.

\section{Motivating Example}\label{sec:motivating_ex}

We illustrate the implications of survey-aware methods using NHANES 2021--2023. NHANES uses stratified and clustered sampling with intentional oversampling of older adults and minority groups to support subgroup estimation.

\begin{figure}[!htb]
\centering
\includegraphics[width=0.90\linewidth]{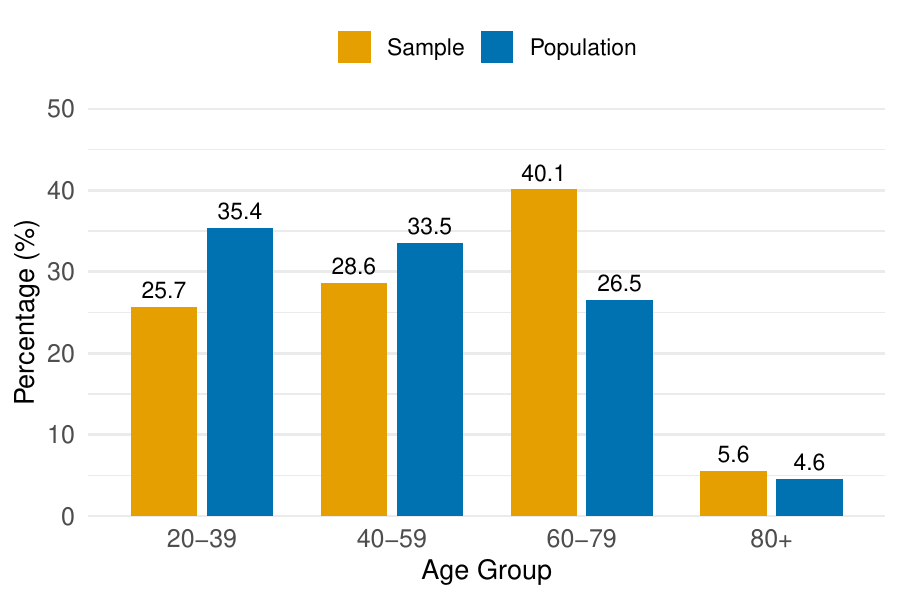} \hfil
\caption{Sample composition vs.\ population estimates by age group (NHANES 2021--2023). Older adults are oversampled to enable precise subgroup estimation.}
\label{fig:exp_composition_age}
\end{figure}

Figure~\ref{fig:exp_composition_age} shows the effect of this design. Older adults are overrepresented in the unweighted sample relative to the U.S. population. This imbalance affects downstream analyses. Unweighted disease prevalence overestimates the population value because high-risk groups are oversampled.

\begin{table}[h]
\small\centering
\caption{Effect of Survey Weights in ML Model Training for Diabetes Prediction}
\label{tab:xgboost_training}
\begin{tabular}{C{1.6cm}C{2.1cm}C{2.1cm}}
\toprule
Training Method & Unweighted AUROC & Weighted AUROC \\
\midrule
Unweighted & 0.782 (0.767--0.795) & 0.809 (0.786--0.827) \\
Weighted & 0.782 (0.768--0.795) & 0.830 (0.815--0.845) \\
\midrule
Difference & 0.000 & +0.021 \\
\bottomrule
\end{tabular}
\end{table}

Table~\ref{tab:xgboost_training} shows the impact on model evaluation. We trained XGBoost classifiers to predict diabetes, with and without survey weights in the loss function. Under unweighted evaluation, both models achieve nearly identical AUROC. Under survey-weighted evaluation, the weighted-trained model has a higher point estimate, although the confidence intervals overlap.

It is important to note that the goal is not to show that weighted methods yield better performance.
Instead, the gap between weighted and unweighted metrics reflects differences between the observed sample and the target population. Unweighted metrics estimate performance in the sample. Weighted metrics estimate performance in the source population defined by the survey design.
When this gap is large, conclusions from standard ML evaluation may not generalize to the population of interest.
These metrics target different estimands, and the appropriate choice depends on the inferential goal.

These results motivate the need for guidance on when and how to incorporate survey design into ML pipelines. The following sections present a nine-step guideline that addresses survey design across the ML lifecycle. Full experimental details and additional results appear in Appendix~\ref{sec:appendix_empirical}.

\section{Related Work}\label{sec:related_work}

\paragraph{Survey Methodology in Statistical Inference.}
Valid inference from complex surveys requires explicit integration of stratification, clustering, and unequal selection probabilities \citep{valliant_practical_2013, heeringa_applied_2017}. Most ML evaluation relies on the i.i.d.\ assumption \citep{j_deng_imagenet_2009, lakshminarayanan_simple_2017, wang_glue_2018}, which obscures data generation processes where sampling design influences validity more than sample size \citep{xiao-li_meng_statistical_2018}. Real-world health data rarely follow simple random sampling and reflect selection mechanisms and measurement constraints that induce structural dependence. Large biobanks and electronic health records show volunteer bias, temporal correlation, and geographic clustering \citep{van_alten_reweighting_2024, johnson_mimic-iii_2016}.

\paragraph{Survey-naive Practice in Health ML.}
Empirical evidence shows that many health ML studies ignore survey structure \citep{west_how_2016, macnell_implementing_2023}. Survey-naive models are often optimized and evaluated with metrics that emphasize sample-level performance rather than population validity \citep{ahn_exploring_2024, chowdhury_investigation_2024}. Unweighted optimization yields parameter estimates driven by sampling artifacts rather than population physiology \citep{pfeffermann_role_1993, macnell_implementing_2023}. Variance estimators that assume independence underestimate uncertainty in clustered data, leading to narrow confidence intervals and inflated false-positive rates \citep{binder_variances_1983, stockwell_estimating_2016}. Fairness assessments that ignore population weights distort algorithmic bias and can obscure disparities affecting underrepresented subpopulations \citep{obermeyer_dissecting_2019, mehrabi_survey_2021, schuch_fairness_2023}.

\paragraph{Alternative Views on Survey Weights.}
Recent studies report comparable performance of unweighted ML for certain tasks \citep{si_machine_2025} or omit survey design considerations without implementation details \citep{chen_identifying_2024}. 
These positions apply when the objective is sample-specific prediction or when the sampling mechanism is ignorable. Such conditions are difficult to verify in practice, particularly for black-box models. Comparable predictive accuracy does not guarantee unbiased population-level estimates \citep{macnell_implementing_2023}.
We frame SaML as a robustness check rather than a strict requirement. When weighted and unweighted results agree, simpler pipelines suffice; divergence signals the need for survey-aware methods.

\paragraph{Gap Between Methods and Practice.}
Despite more than a decade of survey-aware methodological development, adoption remains limited. Audits of health services research show that most studies using complex survey data ignore design features \citep{ridgeway_propensity_2015, west_how_2016}. This pattern persists in algorithmic fairness research \citep{mhasawade_machine_2021}. The gap between methodological advances and standard practice motivates the need for accessible guidelines that clarify when and how to apply survey-aware methods.
In this paper, the proposed guideline covers methods commonly used in survey data analysis but does not address every possible scenario.


\begin{table*}[!t]
    \small\centering
    \caption{Survey-aware Machine Learning (SaML) Guideline: A structured approach for applying ML methods to complex survey data. Each step identifies where standard ML assumptions fail and specifies the survey-aware modification required for valid population inference.
    If a pipeline uses any standard approach listed, the corresponding survey-aware modification is recommended.}
    \label{tab:saml_pipeline}
    \begin{tabular}{@{}c P{0.32\linewidth} P{0.16\linewidth} P{0.42\linewidth}@{}}
    \toprule
    & \textbf{Standard ML Approach} & \textbf{Common Pitfall} & \textbf{Survey-aware Modification} \\
    \midrule
    \multicolumn{4}{l}{\textit{\textbf{A. Data \& Model Development}}} \\[1pt]
    S1 & Standard techniques for missingness and feature engineering. EDA relies on unweighted sample statistics. & Biased descriptive statistics. & Preserve all survey design metadata. EDA must use weighted statistics. Imputation must account for the survey design (e.g., conditioning on strata and clusters). \\ \midrule
    S2 &  Data are randomly split into train/validation/test sets, assuming exchangeability. & Data leakage. & Splitting must respect cluster structure (PSUs) to prevent data leakage. Survey-aware cross-validation is required. \\ \midrule
    S3 & Models optimize standard loss functions, treating each observation equally. & Favoring oversampled subpopulations. & Weighted loss functions are required to optimize for population-level performance, ensuring estimates reflect population characteristics rather than sampling artifacts. \\
    \midrule 
    \multicolumn{4}{l}{\textit{\textbf{B. Evaluation \& Selection}}} \\[1pt]
    S4 & Model selection is guided by performance on an unweighted validation set. & Suboptimal model selection. & Tuning must use weighted metrics (e.g., design-adjusted AIC/BIC) on a survey-aware validation split. \\ \midrule
    S5 & Performance is assessed using standard metrics on the raw test set. & Inaccurate performance estimates. & Weighted evaluation metrics (e.g., weighted AUC, sensitivity, specificity) are essential for unbiased population-level performance estimates. \\ \midrule
    S6 & Uncertainty is estimated assuming i.i.d.\ data (e.g., standard bootstrap).& Underestimating errors. & Variance must be estimated using design-based methods (e.g., Taylor linearization, jackknife, BRR with replicate weights). \\
    \midrule 
    \multicolumn{4}{l}{\textit{\textbf{C. Generalization \& Deployment}}} \\[1pt]
    S7 & Generalizability is assumed or tested informally without accounting for sampling differences.& Unrepresentative generalization. & Performance on external populations requires survey-aware transportability or recalibration methods. \\ \midrule
    S8 & Inference is informal or based on incorrect variance estimates. & Invalid hypothesis tests. & Design-based variance estimation is required for hypothesis testing. Causal effect estimation must integrate survey weights with propensity score methods. \\ \midrule
    S9 & Models are deployed assuming new data follows training distribution.& Model miscalibration. & Predictions may require poststratification or recalibration to reflect the structure of the target population. \\
    \bottomrule
    \end{tabular}
\end{table*}

\section{SaML Guideline}\label{sec:guide}

Table~\ref{tab:saml_pipeline} presents the SaML guideline, organized into three categories (A--C) with nine methodological steps (S1--S9). Category A (Data and Model Development) specifies how data preparation and model training should respect survey structure. Category B (Evaluation and Selection) ensures that model comparison and performance assessment reflect population-level validity rather than sample artifacts. Category C (Generalization and Deployment) addresses how models extend beyond the training survey to external populations and real-world use.

For each step, we contrast the standard ML approach, which assumes independent and identically distributed (i.i.d.) observations, with the survey-aware modification required for valid population-level inference. Key survey design elements include Primary Sampling Units (PSUs, such as counties or hospitals), stratification variables, and sampling weights.
Based on the scoping review in Section~\ref{sec:review}, we provide task-specific guidance on which steps are essential, recommended, or unnecessary for common analytical objectives.

\paragraph{Distinction from Prior Work.}
Existing tutorials on survey analysis~\citep{lohr_sampling_2021,heeringa_applied_2017} are written for statisticians and do not address machine learning procedures such as cross-validation, hyperparameter tuning, or gradient-based optimization. Reproducibility guidelines in machine learning~\citep{kapoor_leakage_2023} focus on experimental practice but do not consider survey design. Reporting standards address parts of this gap. 
TRIPOD (Transparent Reporting of a multivariable prediction model for Individual Prognosis Or Diagnosis)~\citep{collins_transparent_2015} promotes transparent reporting of prediction models but omits survey design. PRICSSA (Preferred Reporting Items for Complex Sample Survey Analysis)~\citep{seidenberg_preferred_2023} specifies reporting requirements for complex surveys, including weighting and variance estimation, but focuses on what to report rather than how to implement survey-aware pipelines. SaML complements these efforts by providing methodological guidance.

\subsection{Data \& Model Development}

\paragraph{S1: Data Preprocessing \& EDA.}
Valid inference requires preservation of design metadata. Unlike standard pipelines that strip auxiliary columns, SaML requires strata, PSUs, and sampling weights to be retained throughout preprocessing. This applies to missing data handling such as from non-response. 
\citet{reiter_importance_2006} showed that ignoring survey design in multiple imputation leads to biased estimates and proposed hierarchical imputation models that preserve cluster structure. Exploratory Data Analysis (EDA) must use weighted descriptive statistics to represent population characteristics rather than sample artifacts.
When sampling weights vary substantially, weight trimming or winsorization can stabilize estimates while introducing a small bias.

\paragraph{S2: Data Splitting \& Validation.}
A critical failure mode in survey-naive ML is data leakage caused by random splitting. When observations from the same PSU such as the same hospital or neighborhood appear in both training and test sets, the model may memorize cluster-specific noise rather than learning generalizable patterns. To prevent this, \citet{wieczorek_k_2022} developed survey-aware cross-validation strategies where folds respect the cluster structure. This PSU-level splitting produces error estimates that are more reliable than random splitting.

\paragraph{S3: Model Training.}
Standard loss functions minimize empirical risk over the sample and favor over-sampled subpopulations. SaML mandates the use of weighted loss functions to ensure population-level optimization. \citet{toth_building_2011} established the theoretical basis for probability-weighted recursive partitioning to ensure design consistency in tree-based models. Similarly, \citet{lumley_aic_2015} demonstrated that loss functions must incorporate sampling weights to approximate the population risk, a principle \citet{dagdoug_model-assisted_2023} extended to high-dimensional penalized regression. Empirically, \citet{macnell_implementing_2023} showed that weighted XGBoost training yields different predictive distributions compared to unweighted alternatives, confirming that design-blind training alters the learned decision boundaries.

\subsection{Evaluation \& Selection}

\paragraph{S4: Hyperparameter Tuning.}
Model selection processes are susceptible to design-induced bias. Hyperparameters selected to minimize unweighted cross-validation error do not necessarily minimize population-level prediction error. \citet{wieczorek_k_2022} demonstrated that valid tuning requires weighted performance metrics evaluated on survey-aware splits.

\paragraph{S5: Performance Evaluation.}
Reporting standard metrics on survey data is misleading because they target a sample-level estimand rather than the population.
\citet{yao_estimation_2015} developed weighted AUROC estimators equivalent to weighted Mann-Whitney U-statistics, while \citet{iparragirre_estimation_2023} extended this approach with bootstrap-based variance estimation. \citet{holbrook_estimating_2020} derived Horvitz-Thompson prediction error estimators, and \citet{wadekar_evaluating_2024} generalized these concepts to weighted estimators for sensitivity and specificity. 
For example, weighted AUROC assigns each concordant pair a weight proportional to the inverse selection probabilities of the two observations.

\paragraph{S6: Uncertainty Quantification.}
Standard ML or bootstrap methods assume i.i.d.\ observations and underestimate variance when applied to clustered data~\citep{binder_variances_1983}. 
For non-linear metrics such as AUROC, \citet{yao_estimation_2015} implemented design-based resampling methods including the jackknife and Balanced Repeated Replication (BRR), a variance estimation technique using half-sample replicates.
\citet{wadekar_evaluating_2024} extended Taylor series linearization to provide valid confidence intervals for binary classification metrics.
In general, Taylor linearization suits smooth statistics such as means and regression coefficients, where analytic derivatives are available. Resampling methods (such as jackknife, BRR, and bootstrap) suit non-smooth or composite statistics. 

\subsection{Generalization \& Deployment}

\paragraph{S7: External Validity.}
Models trained on one survey may not generalize to other populations.
\citet{dong_using_2020} show that incorporating survey weights at both estimation and outcome analysis stages is required for valid population-level inference. \citet{degtiar_review_2023} review methods for assessing generalizability and transportability and distinguish extension to the source population from transport to new target populations.
When survey design metadata are unavailable in the deployment setting, distribution shift between the training and target populations may instead be addressed with domain adaptation methods that learn population-invariant representations \citep{oh_multi-view_2025}.

\paragraph{S8: Statistical Inference.}
Valid hypothesis testing requires correct variance estimates.
\citet{lumley_aic_2015} developed design-based Wald statistics to enable rigorous model comparison under complex sampling designs. \citet{dagdoug_model-assisted_2023} extended these inference capabilities to high-dimensional settings, allowing valid variable selection and significance testing in penalized regression models. When ML is used for causal estimation such as Population Average Treatment Effects (PATE), survey weights must be integrated into the estimators. \citet{ridgeway_propensity_2015} showed that ignoring weights biases PATE estimates by failing to reconstruct the population, while \citet{lenis_its_2019} clarified that weights are required during outcome analysis even if omitted during propensity score estimation.

\paragraph{S9: Deployment \& Calibration.}
Deployed models may require calibration to match the target population structure \citep{niculescu-mizil_predicting_2005}, which may differ from the training survey calibration. \citet{zhang_multilevel_2014} demonstrated Multilevel Regression with Poststratification (MRP) for generating census-aligned small-area estimates. \citet{macnell_implementing_2023} noted that post hoc weighted evaluation can partially correct biases from unweighted training but does not replace a survey-aware pipeline.
When models are applied across survey cycles, temporal shifts in population composition can further degrade calibration \citep{meertens_improving_2022,falasinnu_problem_2025}.

\section{Scoping Review of SaML}\label{sec:review}

\subsection{Approach}

This scoping review does not aim to catalog all studies that apply machine learning to survey data nor circumscribe the possibilities of survey-aware architectures. 
Instead, it identifies methodological contributions that inform a survey-aware ML framework. Following \citet{arksey_scoping_2005}, we searched PubMed, Google Scholar, and statistical methodology journals for papers published between 2000 and 2025. Search terms included survey weights, complex survey, sampling design, and stratified sampling, combined with machine learning, prediction, classification, cross-validation, and weighted estimation.
Please refer to Appendix~\ref{sec:appendix_paper_summaries} for details.

Initial searches showed that most ML studies using survey data ignore survey design and apply standard algorithms without sampling weights or adjustment for stratification and clustering. To identify studies with methodological contributions, we combined database searches with backward and forward citation tracking from key papers \citep{wohlin_guidelines_2014, tricco_scoping_2016}. Three reviewers independently screened titles and abstracts, and resolved discrepancies through discussion.

We included papers that explicitly addressed at least one survey design element, such as weights, strata, or PSUs, in ML training, evaluation, or inference, and proposed methods applicable to health research. We excluded papers that used survey data only as a data source without methodological contribution to survey-aware inference. This process identified 16 papers covering tree-based methods, cross-validation, performance evaluation, variance estimation, and statistical inference.

\begin{table*}[!t]
\small\centering
\caption{Key Methodology Papers for Survey-aware Machine Learning.}
\label{tab:saml-papers}
\begin{tabular}{lllll}
\toprule
\textbf{Citation} & \textbf{Domain} & \textbf{Survey Data} & \textbf{Task} & \textbf{Model} \\
\midrule
\citet{reiter_importance_2006} & Health & NHANES & Imputation & MI \\
\citet{hedt_health_2011} & HIV/Infectious & ANC Sentinel & Estimation & MLE \\
\citet{zhang_multilevel_2014} & Respiratory & BRFSS & SAE & MRP \\
\citet{lumley_aic_2015} & Cardiovascular & NHANES & Selection & LR \\
\citet{yao_estimation_2015} & Health & HHANES & ROC/AUC & Nonparametric \\
\citet{holbrook_estimating_2020} & Renal & NHANES III & Prediction & GLM, kNN \\
\citet{ellis_using_2021} & Audiology & NHANES & Classification & RF, SVM, kNN \\
\citet{wieczorek_k_2022} & Reprod.\ Health & NSFG & CV & LASSO, Spline \\
\citet{macnell_implementing_2023} & Mortality & NHANES III & Classification & XGBoost \\
\citet{wieczorek_design-based_2023} & Health Expend. & MEPS, API & Prediction & Conformal, LR \\
\citet{kalpourtzi_handling_2024} & Cardiovascular & EMENO & Imputation & MI, IPW \\
\citet{van_alten_reweighting_2024} & Epidemiology & UK Biobank & Bias Correction & IPW, LASSO \\
\citet{wadekar_evaluating_2024} & Mental Health & NSDUH, NCS-R & Evaluation & LR, RF \\
\citet{bhaduri_review_2025} & Methodological & Simulated & Methods Review & CART, RF, GB, BART \\
\citet{falasinnu_problem_2025} & Chronic Pain & NHIS & Prediction & LASSO, RF \\
\citet{si_machine_2025} & Orthopedic & NHANES & Classification & RF, SVM, DNN \\
\bottomrule
\end{tabular}
\end{table*}

\newcommand{\cmark}{\cellcolor{green!30}$\checkmark$}
\newcommand{\tmark}{\cellcolor{yellow!30}$\triangle$}
\newcommand{\nmark}{\cellcolor{gray!20}\textendash}

\begin{table*}[!t]
\small\centering
\caption{SaML 9-Step Assessment of Key Methodology Papers. Rating levels: 
$\checkmark$ (green) = Survey-aware methodology applied; 
$\triangle$ (yellow) = Acknowledged but not addressed; 
\textendash\ (gray) = Not applicable.}
\label{tab:saml-assessment}
\begin{tabular}{l C{0.7cm}C{0.7cm}C{0.7cm}C{0.7cm}C{0.7cm}C{0.7cm}C{0.7cm}C{0.7cm}C{0.7cm}}
\toprule
\textbf{Citation} & 
{\textbf{S1}} & {\textbf{S2}} & {\textbf{S3}} & 
{\textbf{S4}} & {\textbf{S5}} & {\textbf{S6}} & 
{\textbf{S7}} & {\textbf{S8}} & {\textbf{S9}} \\
\midrule
\citet{reiter_importance_2006} & \cmark & \nmark & \cmark & \nmark & \tmark & \cmark & \tmark & \tmark & \tmark \\
\citet{hedt_health_2011} & \tmark & \nmark & \cmark & \cmark & \cmark & \cmark & \cmark & \tmark & \tmark \\
\citet{zhang_multilevel_2014} & \cmark & \nmark & \cmark & \nmark & \tmark & \cmark & \cmark & \tmark & \cmark \\
\citet{lumley_aic_2015} & \tmark & \nmark & \tmark & \cmark & \tmark & \tmark & \nmark & \cmark & \tmark \\
\citet{yao_estimation_2015} & \tmark & \nmark & \cmark & \nmark & \cmark & \cmark & \tmark & \tmark & \tmark \\
\citet{holbrook_estimating_2020} & \tmark & \nmark & \cmark & \cmark & \cmark & \cmark & \cmark & \tmark & \tmark \\
\citet{ellis_using_2021} & \cmark & \tmark & \nmark & \nmark & \tmark & \nmark & \tmark & \cmark & \tmark \\
\citet{wieczorek_k_2022} & \tmark & \cmark & \tmark & \cmark & \cmark & \tmark & \tmark & \nmark & \tmark \\
\citet{macnell_implementing_2023} & \tmark & \tmark & \cmark & \tmark & \cmark & \tmark & \cmark & \nmark & \tmark \\
\citet{wieczorek_design-based_2023} & \tmark & \cmark & \tmark & \nmark & \cmark & \cmark & \cmark & \tmark & \tmark \\
\citet{kalpourtzi_handling_2024} & \cmark & \nmark & \cmark & \tmark & \cmark & \cmark & \tmark & \cmark & \tmark \\
\citet{van_alten_reweighting_2024} & \cmark & \nmark & \cmark & \tmark & \tmark & \tmark & \cmark & \tmark & \cmark \\
\citet{wadekar_evaluating_2024} & \tmark & \tmark & \tmark & \nmark & \cmark & \tmark & \cmark & \nmark & \tmark \\
\citet{bhaduri_review_2025} & \cmark & \cmark & \cmark & \tmark & \cmark & \cmark & \tmark & \tmark & \tmark \\
\citet{falasinnu_problem_2025} & \tmark & \nmark & \cmark & \tmark & \cmark & \cmark & \cmark & \nmark & \tmark \\
\citet{si_machine_2025} & \tmark & \nmark & \tmark & \tmark & \tmark & \nmark & \tmark & \nmark & \nmark \\
\bottomrule
\end{tabular}
\end{table*}

\begin{table*}[!t]
\small\centering
\caption{SaML Steps by Machine Learning Task ($\bullet$ = essential, $\circ$ = recommended).}
\label{tab:task_steps}
\begin{tabular}{@{}l *{9}{C{0.9cm}} @{}}
\toprule
& \multicolumn{3}{c}{\textbf{A. Data \& Model}} & \multicolumn{3}{c}{\textbf{B. Evaluation}} & \multicolumn{3}{c}{\textbf{C. Generalization}} \\
\cmidrule(lr){2-4} \cmidrule(lr){5-7} \cmidrule(lr){8-10}
\textbf{Task} & S1 & S2 & S3 & S4 & S5 & S6 & S7 & S8 & S9 \\
\midrule
Descriptive / EDA           & $\bullet$ & --        & --        & --        & --        & $\circ$   & --        & --        & -- \\
Prediction                  & $\bullet$ & $\bullet$ & $\bullet$ & $\circ$   & $\bullet$ & $\circ$   & --        & --        & $\circ$ \\
Risk Factor Identification  & $\bullet$ & $\circ$   & $\bullet$ & $\bullet$ & $\circ$   & $\circ$   & --        & $\bullet$ & -- \\
Small Area Estimation       & $\circ$   & --        & $\bullet$ & --        & --        & --        & $\bullet$ & --        & $\circ$ \\
\bottomrule
\end{tabular}
\end{table*}

\subsection{Analysis}

We assessed each paper using a three-level rating system: \textbf{Applied} ({$\checkmark$}) indicates survey-aware methodology was implemented, \textbf{Acknowledged} ({$\triangle$}) indicates the issue was recognized but not addressed, and \textbf{Not Applicable} ({\textendash}) indicates the step was outside the paper's scope. Three researchers independently assessed each paper; discrepancies were resolved through discussion with reference to textual evidence. 
Detailed rating criteria and paper summaries are provided in the Appendix~\ref{sec:appendix_paper_summaries}.

Table~\ref{tab:saml-papers} summarizes the selected papers. These 16 studies do not represent a comprehensive census of the field. Each paper contributes a methodological building block to one or more steps of the framework.
The papers span diverse healthcare domains, including cardiovascular, respiratory, mental health, and chronic pain. They use nationally representative surveys primarily from the United States. The National Health and Nutrition Examination Survey (NHANES) and its variants appear most frequently.
Other major surveys include the Behavioral Risk Factor Surveillance System (BRFSS), National Health Interview Survey (NHIS), Medical Expenditure Panel Survey (MEPS), and National Survey of Family Growth (NSFG). International surveys include the National Survey of Morbidity and Risk Factors (EMENO, Greece) and UK Biobank. 

Notably, \citet{bhaduri_review_2025} reviews tree-based methods using simulation studies under complex survey designs, comparing weighted and unweighted ensembles and showing that survey weights reduce bias and mean-squared error.

Table~\ref{tab:saml-assessment} presents the assessment results. Steps S3 (Model Training) and S5 (Performance Evaluation) receive the most Applied ($\checkmark$) ratings, reflecting concentrated methodological effort.
S6 (Uncertainty Quantification) receives comparable attention, building on established variance estimation techniques in survey statistics.

In contrast, earlier and later pipeline steps remain underexplored. S9 (Deployment and Calibration) has only two Applied ratings, indicating a critical gap in translating survey-aware models into practice. S2 (Data Splitting) also receives limited attention despite its role in preventing data leakage. Only \citet{wieczorek_k_2022, wieczorek_design-based_2023} and \citet{bhaduri_review_2025} provide Applied contributions. S4 (Hyperparameter Tuning) is addressed by four papers but remains less developed than training and evaluation.
 
The literature is fragmented, and most papers address only a few steps. \citet{bhaduri_review_2025} covers the broadest range, but no single work addresses all nine steps.
This fragmentation motivated our focus on developing a unifying guideline rather than conducting an exhaustive systematic review.

The 16 methodology papers in Table~\ref{tab:saml-assessment} provide building blocks for individual steps, but no single work covers the full lifecycle required in ML applications. \citet{west_how_2016} show that most secondary analyses of complex survey data ignore design features. Recent ML studies on NHANES and BRFSS show that this pattern persists \citep{ahn_exploring_2024, chowdhury_investigation_2024}. Table~\ref{tab:task_steps} maps these fragmented methods into task-specific guidance, specifying which steps are essential or recommended for each objective. A systematic review of recent health ML studies would further strengthen the empirical basis for these guidelines.
Even before such an audit is available, the task-specific mapping in Table~\ref{tab:task_steps} provides immediately usable guidance for ML studies on complex health surveys.

\subsection{Task-Specific Use of the Framework}\label{sec:task}
Although the SaML guideline includes nine steps, not every step is required for all machine learning tasks. Table~\ref{tab:task_steps} summarizes the required steps for four common survey-based ML tasks.
Furthermore, as shown in Appendix~\ref{sec:appendix_empirical}, the impact of each step can vary. Empirical results indicate that the choice of evaluation metric (S5) produces larger and more consistent differences than the choice of training method (S3).
In this work, we summarize the recommended steps for four survey-based ML tasks in Table~\ref{tab:task_steps}. An objective of future work is to assess and quantify the relative impact of each step to ascertain the extent of their importance for each task.

\paragraph{Descriptive analysis and EDA.}
Weighted preprocessing (S1) is essential. Uncertainty quantification (S6) is recommended when reporting confidence intervals. Conventional statistical research focuses on this task without using ML models.
\citet{reiter_importance_2006} develop survey-aware multiple imputation methods, while \citet{van_alten_reweighting_2024} construct inverse probability weights to correct volunteer bias in UK Biobank.
Prior to imputation, missingness and its mechanisms can also be characterized while accounting for survey design to understand patterns and variance in non-response.

\paragraph{Prediction.}
ML-based prediction tasks require the most extensive coverage. These tasks rely on weighted preprocessing (S1), survey-aware data splitting (S2), weighted model training (S3), and weighted performance evaluation (S5). Hyperparameter tuning with design-adjusted criteria (S4) and deployment calibration (S9) are recommended but less often implemented.
Most surveyed papers focus on prediction, with contributions spanning model training \citep{macnell_implementing_2023,falasinnu_problem_2025} and evaluation metrics development \citep{yao_estimation_2015,wadekar_evaluating_2024}.

\paragraph{Risk factor identification.}
This task differs from prediction by prioritizing survey-aware model selection (S4) and statistical inference (S8) for valid hypothesis testing. Performance evaluation (S5) plays a secondary role.
\citet{lumley_aic_2015} derive design-adjusted AIC and BIC criteria for variable selection in complex surveys.

\paragraph{Small area estimation.}
This task uniquely requires external validity methods (S7) for poststratification to target populations. Deployment calibration (S9) is recommended to generate subpopulation estimates.
\citet{hedt_health_2011} develop design-based estimation for HIV prevalence from sentinel surveillance, and \citet{zhang_multilevel_2014} apply multilevel regression with poststratification (MRP) for county-level COPD prevalence.

This task-specific perspective explains the patterns observed in the literature assessment (Table~\ref{tab:saml-assessment}). Steps such as S2, S7, S8, and S9 appear less frequently not due to limited importance, but because they apply only to specific task types.

\section{Practical Implications}\label{sec:implication}

\paragraph{Reproducibility in Automated Discovery.}
AI systems that automate hypothesis generation \citep{wang_scientific_2023} inherit the biases of their training pipelines. When these pipelines ignore survey design, downstream conclusions may not replicate.
Survey-naive pipelines may produce internally consistent results that fail to replicate when applied to new samples drawn from the same population, undermining the scientific credibility of automated discovery systems \citep{subbaswamy_development_2020}.

Reproducibility remains a challenge in ML-based science, with data leakage a common cause \citep{kapoor_leakage_2023}. In complex health surveys, random sample splitting can induce leakage by placing cluster-level structure in both training and test sets. As a result, survey-naive pipelines may produce internally consistent results that fail to replicate in new samples from the same population \citep{subbaswamy_preventing_2019}.
Survey-aware data splitting (S2) avoids this issue by partitioning at the PSU level, which enforces design-consistent independence between training and evaluation sets. When combined with weighted performance evaluation (S5), this strategy yields performance estimates that reflect generalization to new sampling units rather than memorization of design artifacts.

\paragraph{Algorithmic Fairness as Health Equity.}
In public health applications, fairness requires accurate representation of population-level disease burden rather than sample-level frequencies. Fairness audits based on unweighted samples overemphasize over-sampled subgroups and obscure errors affecting underrepresented populations \citep{schuch_fairness_2023}. For example, a model evaluated on NHANES without accounting for oversampling of minority groups may appear equitable while systematically underperforming for the majority population. SaML addresses this issue through weighted exploratory analysis (S1) to characterize subgroup distributions and weighted performance evaluation (S5) to assess error rates across demographic groups. This alignment ties algorithmic assessment to the demographic distribution of the target population.

This perspective directly links algorithmic fairness to health equity.
Fairness and health equity are connected: biased performance estimates can worsen existing disparities \citep{mehrabi_survey_2021}. Weighted evaluation (S5) and proper variance estimation (S6) align reported metrics with the target population.

\paragraph{SaML as a Robustness Check.}
Unweighted estimation can be appropriate when the data-generating process is fully specified and the sampling mechanism is ignorable. 
Deep learning models rely on implicit assumptions about input-output relationships that are difficult to verify.
Expecting these models to recover complex sampling mechanisms, including stratification, clustering, and unequal selection probabilities, without explicit guidance is risky. SaML serves as a robustness check by comparing weighted and unweighted results across pipeline stages, including model training (S3), performance evaluation (S5), and variance estimation (S6).

Differences between weighted and unweighted estimates can indicate informative sampling or model misspecification \citep{pfeffermann_role_1993}. For example, large gaps between weighted and unweighted AUROC (S5), or substantially wider design-based confidence intervals than naive intervals (S6), signal violations of i.i.d. assumptions. In this setting, SaML provides both correction and diagnosis through systematic comparison at S3, S5, and S6.

\paragraph{Implementation Considerations.}
In current ML workflows, adopting survey-aware methods faces practical barriers.
In R, the \texttt{survey} package\footnote{\url{https://cran.r-project.org/web/packages/survey/}} \citep{lumley_survey_2003,lumley_analysis_2004} supports design-based inference, and recent extensions add survey-aware tree models \citep{toth_building_2011} and cross-validation \citep{wieczorek_k_2022}. 
In contrast, Python libraries such as \texttt{scikit-learn}~\citep{pedregosa_scikit-learn_2011}, \texttt{XGBoost}~\citep{chen_xgboost_2016}, and \texttt{PyTorch}~\citep{paszke_pytorch_2019} lack native support for survey design features.
Practitioners must implement weighted loss functions manually or rely on sample weight options that ignore clustering and stratification. This gap poses a barrier for health data scientists who primarily use Python.

Survey-aware methods introduce modest computational overhead. Design-based variance estimation using Taylor linearization or replicate weights increases evaluation time but does not affect training complexity when only sample weights are used. PSU-level cross-validation can reduce the effective number of folds when clusters are large, which may increase variance in performance estimates. 
This tradeoff reflects generalization uncertainty rather than a methodological artifact.
For large surveys with hundreds of PSUs, computational costs remain manageable on standard hardware. Implementation complexity, rather than computation, remains the main barrier to adoption.

\section{Conclusion}
Machine learning provides powerful tools for analyzing complex health data. Standard practice often ignores sampling weights, stratification, and clustering. 
This omission compromises population-level inference, leading to biased estimates, underestimated variance, and misleading fairness assessments.

We propose Survey-aware Machine Learning (SaML), a nine-step guideline for valid analysis with representative data. SaML requires design-aware data splitting (S2) to prevent leakage, weighted optimization (S3) to address selection bias, and design-based variance estimation (S6) for uncertainty quantification. These steps ensure that model outputs reflect the target population rather than sampling artifacts.
As artificial intelligence systems increasingly automate scientific discovery and clinical decision-making, methodological guardrails become necessary. 
Survey-aware methods should become standard practice when analyzing complex health surveys.

\paragraph{Limitations.}
The scoping review focuses on methodological contributions rather than exhaustive coverage. 
The empirical illustration validates individual steps but does not evaluate the full pipeline jointly. 
Furthermore, practical limitations remain. Highly skewed weights reduce effective sample size. High item non-response can weaken the validity of survey weights. Temporal distribution shift across annual survey cycles also remains largely unaddressed in the current literature.
Lastly, the empirical illustration uses a single survey dataset, with a simplified prediction task. Generalizability of the observed patterns across different survey designs and outcome types requires further validation.

\paragraph{Open Problems.}
Survey-adjusted hyperparameter tuning and deployment calibration lack established procedures and remain major methodological gaps identified by our review. Formal comparison of SaML methods with causal inference frameworks and weighted tree-based approaches remains an open direction.
Validation of the full pipeline as a whole, rather than isolated steps, is a necessary next step. Extending survey-aware principles to non-probability samples such as electronic health records and establishing reporting standards for survey-based ML analyses remain open priorities in our community.

\acks{
The authors thank the researchers whose methodological work on complex survey analysis and machine learning informed this study.

The authors used AI-assisted tools for language editing and formatting review. The authors reviewed and approved all content, analyses, and conclusions.

Dr. YongKyung Oh was supported by the Basic Science Research Program through the National Research Foundation of Korea (NRF) funded by the Ministry of Education (RS-2024-00407852).
}

{\small
\bibliography{references}
}

\clearpage
\newpage
\appendix
\onecolumn

\section{Empirical Illustration}\label{sec:appendix_empirical}
To demonstrate the practical implications of survey-aware methods, we present an empirical analysis using National Health and Nutrition Examination Survey (NHANES) August 2021--August 2023 cycle. This illustration quantifies the differences between standard (unweighted) and survey-aware (weighted) approaches across multiple stages of the machine learning pipeline.

Experiments 1--3 compare weighted and unweighted approaches: 
Experiment 1 (Descriptive Statistics) validates S1, 
Experiment 2 (Model Evaluation) validates S5 and S6, and 
Experiment 3 (Training with Weights) validates S3 and S5. 
Furthermore, Experiment 4 (Cross-Validation Design) examines how PSU-level versus random splitting affects performance estimation, validating S2 and S4.

\paragraph{Data.} NHANES employs a complex, multistage probability sampling design to produce nationally representative estimates of the U.S. civilian non-institutionalized population. The survey design includes stratification (SDMVSTRA), clustering by primary sampling units (SDMVPSU), and differential selection probabilities captured in examination weights (WTMEC2YR). NHANES oversamples older adults, racial and ethnic minorities, and low-income individuals to support subgroup estimation.

For the survey-design, we merged demographic (DEMO\_L), body measurement (BMX\_L), blood pressure (BPXO\_L), and diabetes questionnaire (DIQ\_L) data files. Our analysis sample consisted of adults aged 20 years and older with valid examination weights ($n = 6064$).

\paragraph{Task.} The prediction task was binary classification of diabetes status, defined as self-reported physician diagnosis (DIQ010 = ``Yes''). Among participants with valid diabetes status ($n = 5843$), the unweighted prevalence was 14.9\%, compared to a survey-weighted population estimate of 12.1\%, reflecting the oversampling of older adults with higher disease risk.
Predictors included age (years) and body mass index (kg/m$^2$), two routinely measured clinical variables.

This simple model specification isolates the effects of survey methodology from model complexity. Any differences between weighted and unweighted approaches therefore reflect the survey design rather than model misspecification or feature engineering.

\subsection{Experiment 1: Descriptive Statistics}
\label{subsec:exp1}

The first experiment shows how unweighted sample statistics can diverge from design-weighted population estimates when sampling is non-proportional.

\subsubsection{Continuous Variables}

\begin{figure}[!htb]
\centering
\includegraphics[width=0.21\linewidth]{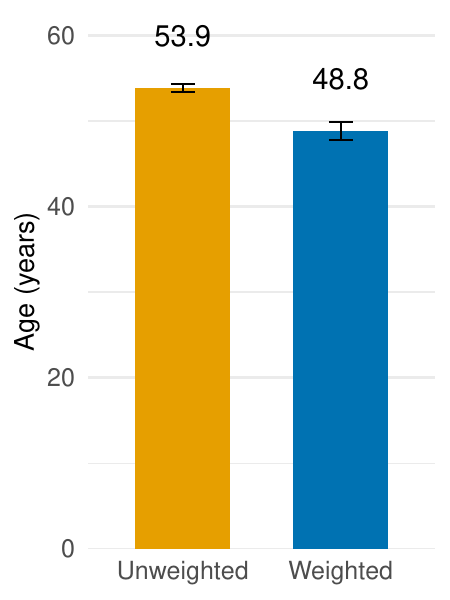} \hfil
\includegraphics[width=0.21\linewidth]{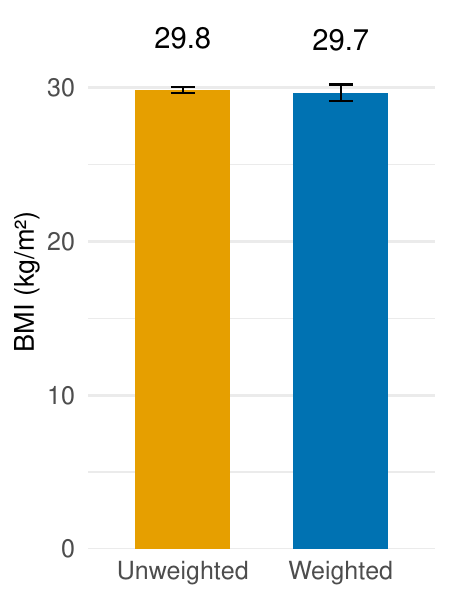} \hfil
\includegraphics[width=0.21\linewidth]{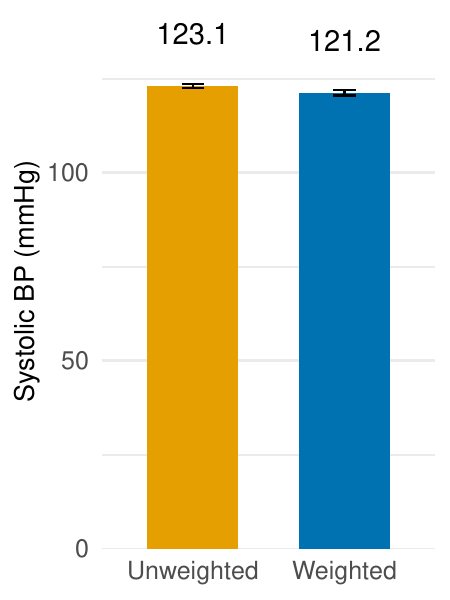} \hfil
\includegraphics[width=0.21\linewidth]{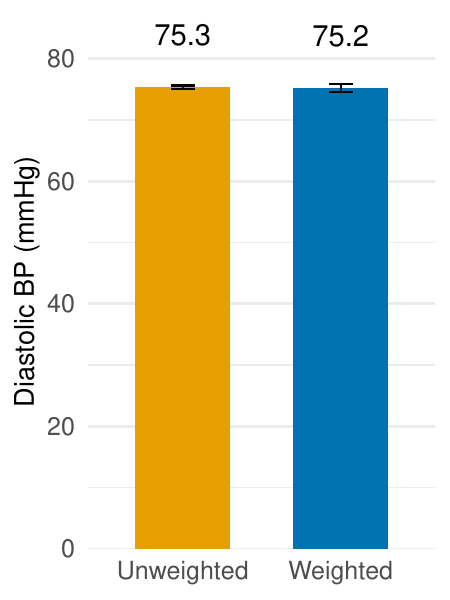}
\caption{Comparison of unweighted and weighted estimates for continuous variables (Age, BMI, Systolic BP, Diastolic BP, from left to right). Error bars indicate 95\% confidence intervals.}
\label{fig:exp_continuous}
\end{figure}

Figure~\ref{fig:exp_continuous} and Table~\ref{tab:weighted_continuous} compare unweighted and weighted means for key health indicators (NHANES 2021--2023, U.S. adults $\geq$ 20 years, $n = 6064$). Unweighted standard errors assume simple random sampling, $\text{SE} = s / \sqrt{n}$, while weighted standard errors account for the complex survey design via Taylor linearization.

\begin{table}[htb]
\small\centering
\caption{Comparison of Unweighted and Weighted Estimates for Continuous Variables} 
\label{tab:weighted_continuous}
\begin{tabular}{lcccccc}
\toprule
\multirow{2.5}{*}{Variable} & \multicolumn{2}{c}{Unweighted} & \multicolumn{2}{c}{Weighted} & \multirow{2.5}{*}{Diff} & \multirow{2.5}{*}{\% Diff} \\
\cmidrule(lr){2-3} \cmidrule(lr){4-5}
 & Mean & SE & Mean & SE & & \\
\midrule
Age (years) & 53.9 & 0.22 & 48.8 & 0.54 & -5.1 & -9.4\% \\
BMI (kg/m$^2$) & 29.8 & 0.10 & 29.7 & 0.27 & -0.1 & -0.5\% \\
Systolic BP (mmHg) & 123.1 & 0.24 & 121.3 & 0.35 & -1.8 & -1.5\% \\
Diastolic BP (mmHg) & 75.3 & 0.15 & 75.2 & 0.31 & -0.1 & -0.2\% \\
\bottomrule
\end{tabular}
\end{table}

The largest difference appears in age. The unweighted sample mean exceeds the weighted population estimate by 5.1 years, reflecting NHANES's intentional oversampling of older adults. This 9.4\% relative difference shows how sample composition can misrepresent the target population.

Figure~\ref{fig:exp_continuous} visualizes these differences. Age shows the largest discrepancy, with the unweighted estimate exceeding the weighted estimate because older adults are oversampled. Body mass index and blood pressure show minimal differences, indicating weaker sensitivity to the sampling design.

\subsubsection{Disease Prevalence}

Table~\ref{tab:weighted_prevalence} shows that unweighted prevalence estimates systematically overestimate population disease burden. Unweighted prevalence exceeds weighted estimates for both diabetes (defined as self-reported physician diagnosis) and hypertension (defined as systolic BP $\geq$ 140 or diastolic BP $\geq$ 90 mmHg). These differences arise because older adults, who have higher disease prevalence, are overrepresented in the unweighted sample.
\begin{table}[htb]
\small\centering
\caption{Comparison of Unweighted and Weighted Prevalence Estimates}
\label{tab:weighted_prevalence}
\begin{tabular}{lccccc}
\toprule
Outcome & $n$ & Unweighted \% (SE) & Weighted \% (SE) & Diff (pp) \\
\midrule
Diabetes & 5843 & 14.9 (0.47) & 12.1 (0.71) & -2.8 \\
Hypertension & 5863 & 20.6 (0.53) & 17.9 (0.95) & -2.7 \\
\bottomrule
\end{tabular}
\end{table}

\begin{figure}[!htb]
\centering
\includegraphics[width=0.48\linewidth]{figs/empirical/fig_exp1c_age.pdf} \hfil
\includegraphics[width=0.48\linewidth]{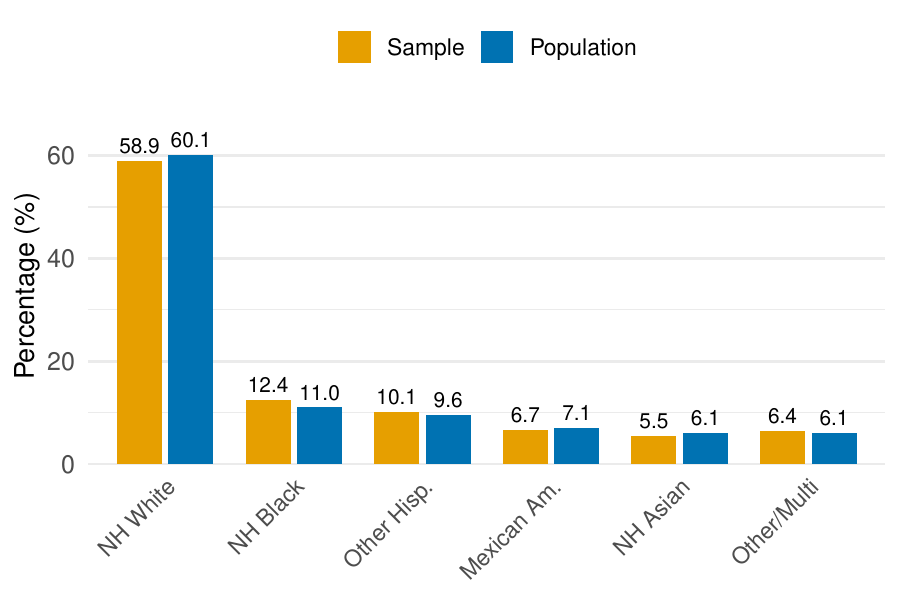}
\caption{Sample composition vs.\ population estimates by age group (left) and race/ethnicity (right).}
\label{fig:exp_composition}
\end{figure}

\subsubsection{Sample Composition by Age Group and Race/Ethnicity}

Figure~\ref{fig:exp_composition} shows these differences of composition in sample and population. Older adults are overrepresented in the unweighted sample relative to the U.S. population, while younger adults are underrepresented. This deliberate oversampling supports precise subgroup estimates but requires weighting for valid population inference. In contrast, race and ethnicity show smaller discrepancies in the NHANES 2021--2023 cycle, indicating more balanced representation in recent survey design, as detailed in Table~\ref{tab:age_representation} and \ref{tab:race_representation}.

\begin{table}[htb]
\small\centering
\caption{Sample Composition vs Population Estimates by Age Group}
\label{tab:age_representation}
\begin{tabular}{lcccc}
\toprule
Age Group & $n$ & Sample \% & Population \% (SE) & Diff (pp) \\
\midrule
20-39 & 1559 & 25.7 & 35.4 (1.23) & +9.7 \\
40-59 & 1733 & 28.6 & 33.5 (0.92) & +4.9 \\
60-79 & 2430 & 40.1 & 26.5 (0.71) & -13.6 \\
80+ & 342 & 5.6 & 4.6 (0.47) & -1.0 \\
\bottomrule
\end{tabular}
\end{table}

\begin{table}[htb]
\small\centering
\caption{Sample Composition vs Population Estimates by Race/Ethnicity}
\label{tab:race_representation}
\begin{tabular}{lcccc}
\toprule
Race/Ethnicity (Abbreviation in Figure~\ref{fig:exp_composition}) & $n$ & Sample \% & Population \% (SE) & Diff (pp) \\
\midrule
Non-Hispanic White (NH White) & 3569 & 58.9 & 60.1 (1.91) & +1.2 \\
Non-Hispanic Black (NH Black) & 753 & 12.4 & 11.0 (1.33) & -1.4 \\
Other Hispanic (Other Hisp.) & 614 & 10.1 & 9.6 (1.32) & -0.5 \\
Mexican American (Mexican Am.) & 405 & 6.7 & 7.1 (1.95) & +0.4 \\
Non-Hispanic Asian (NH Asian) & 335 & 5.5 & 6.1 (1.17) & +0.6 \\
Other/Multi-Racial (Other/Multi) & 388 & 6.4 & 6.1 (0.53) & -0.3 \\
\bottomrule
\end{tabular}
\end{table}

\subsection{Experiment 2: Model Evaluation}
\label{subsec:exp2}

The second experiment examines how evaluation methodology affects reported model performance. We fit a logistic regression model predicting diabetes from age and BMI ($n = 5752$ complete cases, with diabetes prevalence 14.7 \%), and compare unweighted and survey-weighted AUROC estimates.
Table~\ref{tab:weighted_auroc} shows that the same model yields different AUROC values under different evaluation methods. The unweighted and weighted AUROC estimates differ by 0.032, a gap that can affect model selection and deployment decisions.

\begin{table}[htb]
\centering
\caption{Comparison of Unweighted and Survey-Weighted AUROC for Diabetes Prediction} 
\label{tab:weighted_auroc}
\begin{tabular}{lcc}
\toprule
Evaluation Method & AUROC & 95\% CI \\
\midrule
Unweighted & 0.743 & (0.727 -- 0.758) \\
Survey-weighted & 0.775 & (0.749 -- 0.793) \\
\midrule
Difference & +0.032 & \\
\bottomrule
\end{tabular}
\end{table}

The unweighted AUROC uses the standard Mann--Whitney U statistic, with confidence intervals derived from the DeLong method. The survey-weighted AUROC uses a Horvitz--Thompson estimator that weights each concordant pair by inverse selection probabilities \citep{iparragirre_estimation_2023}. Confidence intervals for the weighted AUROC are obtained by bootstrap resampling with 100 replications.

Neither estimate is inherently correct. Each targets a different estimand. The unweighted AUROC measures discrimination in the observed sample, whereas the weighted AUROC measures discrimination in the target population. When sampling probabilities correlate with the outcome, as when older adults are oversampled, these estimates diverge. The appropriate metric depends on the inferential goal, whether sample-level performance or population-level generalization.

\subsection{Experiment 3: Training with Survey Weights}
\label{subsec:exp3}

The third experiment tests whether incorporating survey weights into the training loss affects model performance. We trained XGBoost models with identical hyperparameters (max\_depth=3, learning rate=0.1, 100 boosting rounds), varying only the inclusion of survey weights in the loss function. Uncertainty estimates were obtained via bootstrap resampling (100 replications).

Experiments 2 and 3 address related but distinct questions. Experiment 2 isolates evaluation methodology. Using a fixed logistic regression model, it shows that unweighted and weighted AUROC yield different performance estimates even when the model is unchanged. Experiment 3 adds training methodology as a second factor. Using XGBoost with and without survey-weighted loss, it assesses whether weighted training improves population-level performance. Together, the results indicate that both training and evaluation choices matter, but evaluation methodology produces larger and more consistent differences. When only one survey-aware modification is feasible, weighted evaluation should take priority over weighted training.

Table~\ref{tab:xgboost_auroc_auprc} presents the results. Under unweighted evaluation, both models achieve similar performance regardless of training method. Under weighted evaluation, the weighted-trained model has higher point estimates than the unweighted-trained model, although the confidence intervals overlap. This pattern holds for both AUROC and AUPRC, indicating that incorporating survey weights into the loss function optimizes population-level rather than sample-level performance.

\begin{table}[htb]
\small\centering
\caption{Effect of Survey Weights in XGBoost Training: AUROC and AUPRC}
\label{tab:xgboost_auroc_auprc}
\begin{tabular}{lcccc}
\toprule
\multirow{2.5}{*}{Training Method} & \multicolumn{2}{c}{AUROC} & \multicolumn{2}{c}{AUPRC} \\
\cmidrule(lr){2-3} \cmidrule(lr){4-5}
 & Unweighted & Weighted & Unweighted & Weighted \\
\midrule
Unweighted XGBoost & 0.782 (0.767--0.795) & 0.809 (0.786--0.827) & 0.365 (0.337--0.401) & 0.354 (0.321--0.393) \\
Weighted XGBoost & 0.782 (0.768--0.795) & 0.830 (0.815--0.845) & 0.361 (0.328--0.391) & 0.387 (0.341--0.426) \\
\midrule
Difference & 0.000 & +0.021 & -0.004 & +0.033 \\
\bottomrule
\end{tabular}
\end{table}

\begin{figure}[!htb]
\centering
\includegraphics[width=0.42\linewidth]{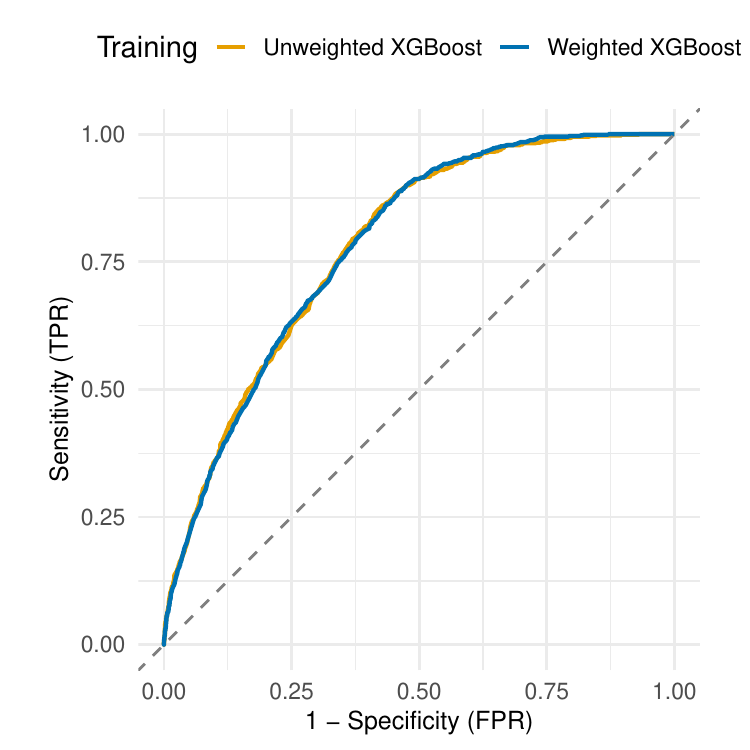} \hfil
\includegraphics[width=0.42\linewidth]{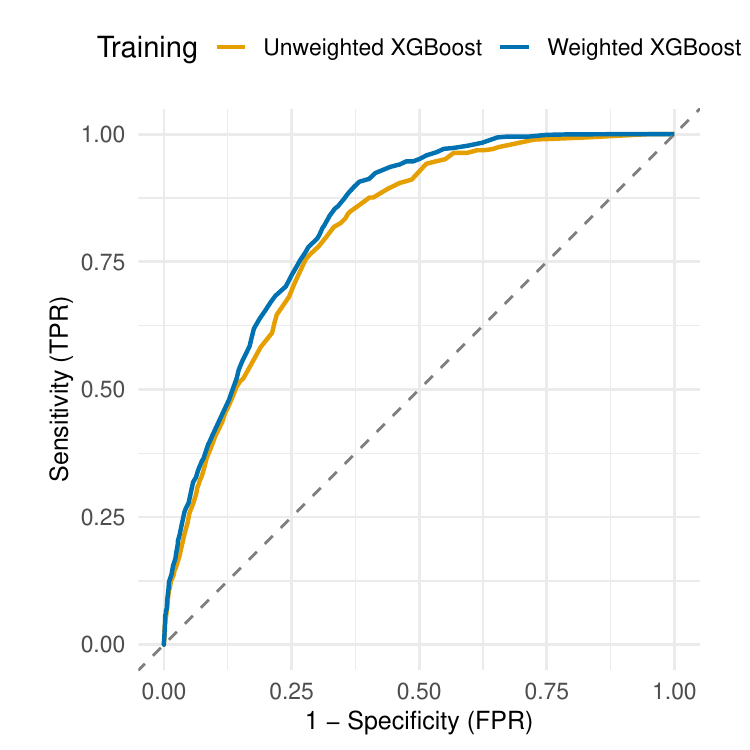}
\caption{ROC curves under standard evaluation (left) and survey-weighted evaluation (right).}
\label{fig:exp_roc}
\end{figure}

Figure~\ref{fig:exp_roc} visualizes these differences using ROC curves. Under standard evaluation (left), the two training methods produce nearly overlapping curves. Under weighted evaluation (right), the weighted-trained model shows improved discrimination, particularly in the high-sensitivity region.

\begin{figure}[!htb]
\centering
\includegraphics[width=0.42\linewidth]{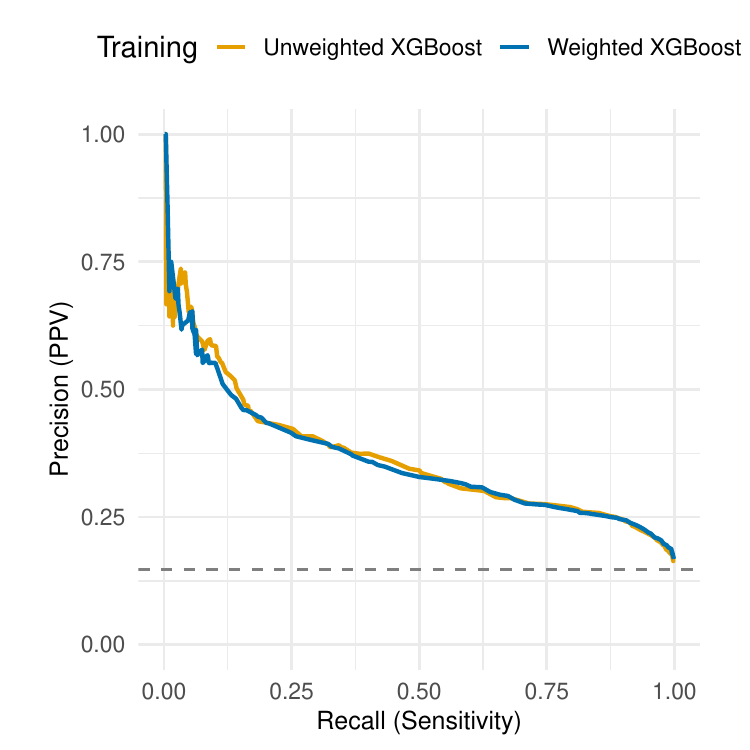} \hfil
\includegraphics[width=0.42\linewidth]{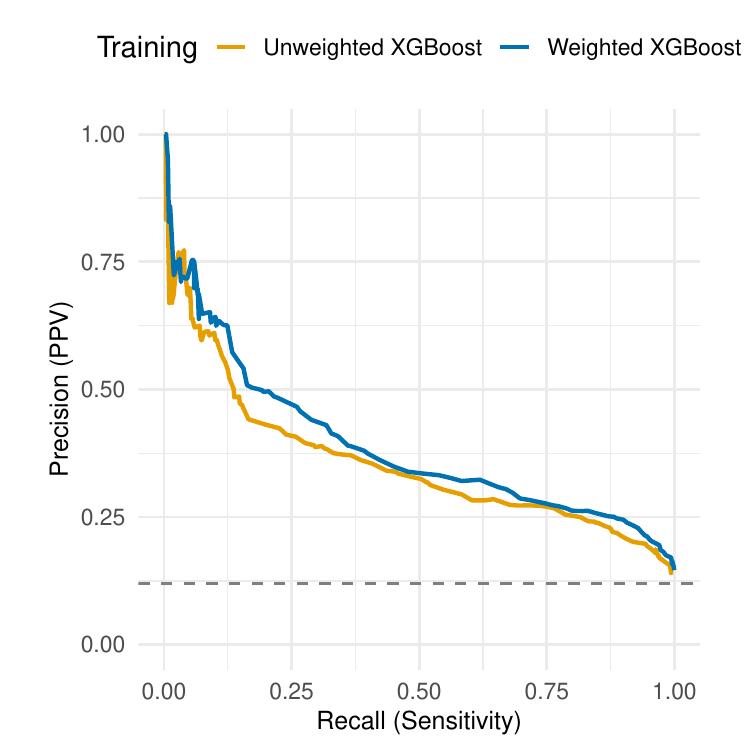}
\caption{Precision-Recall curves under standard evaluation (left) and survey-weighted evaluation (right).}
\label{fig:exp_pr}
\end{figure}

Figure~\ref{fig:exp_pr} shows the corresponding precision--recall curves. The same pattern appears. Training method differences are minimal under standard evaluation but become visible under weighted evaluation, where the weighted-trained model achieves higher precision at comparable recall levels.

\subsection{Experiment 4: Cross-Validation Design}
\label{subsec:exp4}

The fourth experiment examines how cross-validation strategy affects performance estimation. We use the same XGBoost specification as Experiment~3 (max\_depth=3, learning rate=0.1, 100 rounds) in a $2 \times 2 \times 2$ factorial design: CV type (random vs.\ PSU-level), training (unweighted vs.\ weighted loss), and evaluation (unweighted vs.\ weighted AUROC). Each condition is evaluated with 5-fold CV repeated 3 times.

In random CV, observations are assigned to folds independently. In PSU-level CV, fold assignment occurs at the primary sampling unit level and is stratified within each stratum so that every fold contains PSUs from all strata (Algorithm~\ref{alg:psu_cv}). This design preserves the cluster structure and prevents data leakage between training and test sets within the same PSU.

\begin{table}[htb]
\small\centering
\caption{Cross-validation results (mean (SD) over 5-fold CV $\times$ 3 repeats).}
\label{tab:cv_factorial}
\begin{tabular}{llcccc}
\toprule
\multirow{2.5}{*}{CV Type} & \multirow{2.5}{*}{Training} & \multicolumn{2}{c}{AUROC} & \multicolumn{2}{c}{AUPRC} \\ \cmidrule(lr){3-4} \cmidrule(lr){5-6}
& & Unweighted & Weighted & Unweighted & Weighted \\
\midrule
\multirow{2}{*}{Random} 
& Unweighted & 0.739 (0.020) & 0.768 (0.032) & 0.298 (0.026) & 0.284 (0.034) \\
& Weighted   & 0.736 (0.019) & 0.763 (0.033) & 0.288 (0.025) & 0.268 (0.037) \\
\midrule
\multirow{2}{*}{PSU} 
& Unweighted & 0.732 (0.009) & 0.763 (0.008) & 0.283 (0.016) & 0.274 (0.023) \\
& Weighted   & 0.727 (0.011) & 0.756 (0.008) & 0.275 (0.011) & 0.255 (0.015) \\
\bottomrule
\end{tabular}
\end{table}

Table~\ref{tab:cv_factorial} shows a descriptive difference between the two split schemes. For matched training and evaluation settings, PSU-level CV produces slightly lower mean AUROC than random CV.
The AUROC standard deviation is also smaller under PSU-level CV, compared with random CV. 
This stabilization is vital from a public health perspective to reliably assess model fairness for underrepresented groups. 

We interpret this difference cautiously because the PSU design produced only 6 usable folds out of the 15 planned repeat fold combinations after the minimum test size and positive count screen, whereas all 15 random CV folds were retained.
Across both CV types, weighted evaluation increases mean AUROC by about 0.027 to 0.031 relative to unweighted evaluation, but it lowers mean AUPRC in all four training and CV combinations. When the evaluation rule is fixed, weighted training performs slightly worse than unweighted training for both AUROC and AUPRC in this experiment.

In this dataset, the cross-validation design changes both the reported point estimates and their fold-to-fold dispersion. PSU-level CV enforces cluster-separated splits, but with the current sample and positivity thresholds it also drops more folds. We therefore treat Table~\ref{tab:cv_factorial} as a descriptive comparison of two split constructions under the observed survey design, not as a definitive ranking of estimator reliability.

\paragraph{PSU-Level Cross-Validation.} Algorithm~\ref{alg:psu_cv} presents the survey-aware cross-validation procedure used in our empirical illustration. Folds are assigned at the PSU-level within each stratum, following \citet{wieczorek_k_2022}. 
Algorithm 2 defines the survey-weighted AUROC estimator. Specifically, it applies the Horvitz--Thompson estimator \citep{horvitz_generalization_1952} to the Mann--Whitney concordance probability \citep{hanley_meaning_1982} to account for unequal selection probabilities.

\begin{algorithm}[htb]
\small
\caption{Stratified PSU-Level $K$-Fold Cross-Validation}\label{alg:psu_cv}
\DontPrintSemicolon
\SetKwInOut{Input}{Input}
\SetKwInOut{Output}{Output}

\Input{Dataset $\{(\mathbf{x}_i, y_i, w_i, s_i, c_i)\}_{i=1}^{n}$: features, label, weight, stratum, PSU\; \\
Folds $K$, repeats $R$, model class $\mathcal{M}$}
\Output{Performance scores $\mathbf{P} \in \mathbb{R}^{R \times K}$}

\BlankLine
\For{$r \leftarrow 1$ \KwTo $R$}{
  \tcp{Assign folds within each stratum}
  \ForEach{stratum $s$}{
    Randomly assign PSUs in $s$ to folds $1, \ldots, K$\;
  }
  Map each observation $i$ to the fold of its PSU $(s_i, c_i)$\;

  \BlankLine
  \For{$k \leftarrow 1$ \KwTo $K$}{
    \, 
    Train $\hat{m} \leftarrow \mathcal{M}.\text{fit}(\mathbf{X}_{\text{train}},\, \mathbf{y}_{\text{train}},\, \mathbf{w}_{\text{train}})$\; 
    \\ 
    Predict $\hat{p}_i \leftarrow \hat{m}(\mathbf{x}_i)$ for $i$ in fold $k$\; 
    \\ 
    $\mathbf{P}[r, k] \leftarrow \text{WeightedAUROC}(\mathbf{y}_{\text{test}},\, \hat{\mathbf{p}}_{\text{test}},\, \mathbf{w}_{\text{test}})$\;
  }
}
\Return{$\mathbf{P}$}\;
\end{algorithm}

\begin{algorithm}[htb]
\small
\caption{Survey-Weighted AUROC (Horvitz--Thompson Estimator)}\label{alg:weighted_auc}
\DontPrintSemicolon
\SetKwInOut{Input}{Input}
\SetKwInOut{Output}{Output}

\Input{Labels $\mathbf{y}$, predicted scores $\hat{\mathbf{p}}$, sampling weights $\mathbf{w}$}
\Output{$\widehat{\text{AUROC}}_{\text{HT}} \in [0,1]$}

\BlankLine
$\mathcal{P} \leftarrow \{i : y_i = 1\}$, \; $\mathcal{N} \leftarrow \{i : y_i = 0\}$\; \\ 
$W^{+} \leftarrow \textstyle\sum_{i \in \mathcal{P}} w_i$, \; $W^{-} \leftarrow \textstyle\sum_{j \in \mathcal{N}} w_j$\;

\BlankLine
$C \leftarrow 0$\;
\For{$i \in \mathcal{P}$}{
  \For{$j \in \mathcal{N}$}{
    $\omega_{ij} \leftarrow \frac{w_i}{W^{+}} \cdot \frac{w_j}{W^{-}}$\;
    \lIf{$\hat{p}_i > \hat{p}_j$}{$C \leftarrow C + \omega_{ij}$}
    \lElseIf{$\hat{p}_i = \hat{p}_j$}{$C \leftarrow C + \tfrac{1}{2}\omega_{ij}$}
  }
}
\Return{$C$}\;
\end{algorithm}

Based on this approach, all observations from the same PSU remain in the same fold. This prevents within-cluster leakage and preserves cluster integrity. Folds are assigned independently within each stratum so that each fold maintains the stratification structure of the survey design. Both training and evaluation incorporate sampling weights, which targets population-level performance rather than sample-level accuracy.

\subsection{Discussion: Why Differences Matter}\label{subsec:discussion}

The purpose of this empirical illustration is not to claim that weighted methods are universally superior. Instead, it shows that meaningful differences exist between weighted and unweighted approaches, and that ignoring these differences can lead to misleading conclusions about model performance.
When sample and population distributions diverge under complex survey designs, standard ML pipelines may not align with their intended targets. A model optimized on an unweighted sample prioritizes performance on oversampled subgroups, such as older adults in NHANES, rather than the population distribution. Unweighted evaluation metrics similarly reflect discrimination in the observed sample rather than in the target population.

\paragraph{Evaluation methodology matters more than training methodology.}
Across experiments, the choice of evaluation metric, weighted versus unweighted, produces larger and more consistent differences than the choice of training method. This pattern holds for both AUROC and AUPRC, with larger gaps under weighted evaluation for AUPRC. These results suggest that practitioners should prioritize survey-aware evaluation when weighted training is infeasible.

\paragraph{Unweighted and weighted metrics answer different questions.}
Unweighted metrics estimate performance in the observed sample, whereas weighted metrics estimate performance in the target population. Neither metric is universally correct. The appropriate choice depends on the inferential goal. For population-level deployment, weighted evaluation provides the relevant estimand.

\paragraph{Differences should be reported, not ignored.}
Rather than reporting a single metric, practitioners should present both weighted and unweighted results. Large discrepancies indicate sensitivity to sampling design and require explicit discussion. Agreement between metrics suggests robustness to evaluation methodology. The discrepancies observed here, consistent across both AUROC and AUPRC, show that conclusions can depend on methodological choices and that survey design must be addressed explicitly.


\section{Details of Scoping Review}\label{sec:appendix_paper_summaries}

\subsection{Selection Rationale}

This appendix describes the paper selection process. Following the scoping review framework of \citet{arksey_scoping_2005}, the objective was to identify methodological contributions that inform the survey-aware ML framework, not to exhaustively review all ML applications using survey data.

\paragraph{Search Strategy.}
We queried PubMed and Google Scholar using survey design terms (``survey weights,'' ``complex survey,'' ``sampling design,'' ``stratified sampling'') combined with machine learning terms (``machine learning,'' ``prediction,'' ``classification,'' ``cross-validation,'' ``weighted estimation''). 
We also hand-searched multiple journals where survey-aware methods are more commonly published, including but not limited to \textit{Survey Methodology}, \textit{Journal of Survey Statistics and Methodology}, and \textit{Statistics in Medicine}. Searches covered papers published between 2000 and 2025.

\paragraph{Screening.}
Database searches returned many papers applying ML to surveys such as NHANES, BRFSS, UK Biobank, and their case studies. Most treated survey data as a convenience sample without incorporating sampling weights or design adjustments. These papers were excluded because they do not advance survey-aware methodology. 
Given the sparse literature, we supplemented database searches with backward and forward citation tracking from foundational papers \citep{wohlin_guidelines_2014, tricco_scoping_2016} to identify contributions in statistics journals not captured by keyword searches. Three reviewers independently screened titles and abstracts, and resolved discrepancies through discussion.

\paragraph{Inclusion and Exclusion.}
We included papers that (1) explicitly addressed survey design elements (weights, strata, PSUs) in ML training, evaluation, or inference, (2) proposed or evaluated methods rather than applying existing tools, and (3) were applicable to health research.
We excluded papers that used survey data only as a data source or mentioned weights without methodological detail.

\paragraph{Final Selection.}
This process yielded 16 papers spanning model training, cross-validation, performance evaluation, variance estimation, and statistical inference. Although not exhaustive, this set captures the core methodological contributions required for survey-aware ML pipelines.

\subsection{Scoping Review}
\paragraph{Scope of Included Papers.}
The 16 selected papers include both methodological and applied work. Some focus on a single lifecycle step, such as multiple imputation \citep{reiter_importance_2006} or model selection criteria \citep{lumley_aic_2015}. 
Other included papers evaluate ML workflows for survey data, with varying degrees of survey-design integration across training, validation, and evaluation \citep{macnell_implementing_2023,falasinnu_problem_2025,si_machine_2025}.
Section~\ref{sec:related_work} also cites recent studies that train predictive models on nationally representative surveys without accounting for survey design \citep{chen_identifying_2024, ahn_exploring_2024, chowdhury_investigation_2024}, which shows that survey-naive practice persists in applied ML.

\paragraph{Paper Summaries.}
The following summaries describe each paper's contribution to the framework:

\begin{itemize}

\item \citet{reiter_importance_2006} address multiple imputation (MI) for survey data. Conditioning imputation models on strata and cluster indicators prevents bias in design-based inference. Simulation studies using the National Health and Nutrition Examination Survey (NHANES) show that ignoring design features during imputation degrades variance estimation.

\item \citet{hedt_health_2011} combine convenience samples from antenatal care (ANC) sentinel surveillance with probability samples for human immunodeficiency virus (HIV) prevalence estimation. Maximum likelihood estimation (MLE) yields closed-form variance expressions that correct selection bias from convenience sampling.

\item \citet{zhang_multilevel_2014} apply multilevel regression and poststratification (MRP) to small-area estimation (SAE) of chronic obstructive pulmonary disease (COPD) prevalence using the Behavioral Risk Factor Surveillance System (BRFSS). Weight rescaling preserves inference validity in multilevel models, and parametric bootstrap provides uncertainty quantification at the census block level.

\item \citet{lumley_aic_2015} derive design-adjusted Akaike Information Criterion (dAIC) and Bayesian Information Criterion (dBIC) for model selection in complex surveys. Effective sample size corrections and design effects are incorporated through pseudo-likelihood theory, with applications to logistic regression (LR) models for hypertension (NHANES) and esophageal cancer (case-control study).

\item \citet{yao_estimation_2015} develop nonparametric receiver operating characteristic (ROC) curve and area under the curve (AUC) estimators for stratified multistage cluster samples. Variance estimation uses jackknife and balanced repeated replication (BRR), with applications to the Hispanic Health and Nutrition Examination Survey (HHANES).

\item \citet{holbrook_estimating_2020} extend Efron's prediction error estimation to complex surveys. The Horvitz-Thompson-Efron (HTE) estimator is consistent for superpopulation generalization error and equivalent to design-adjusted AIC. Applications use NHANES III with generalized linear models (GLM) and k-nearest neighbors (kNN).

\item \citet{ellis_using_2021} classify hearing loss severity using NHANES audiometric data across multiple survey cycles. A weight combination procedure merges cycles while preserving representativeness. Random forest (RF), support vector machine (SVM), and k-nearest neighbors (kNN) are compared using design-based hypothesis testing.

\item \citet{wieczorek_k_2022} develop survey-aware K-fold cross-validation (CV) for complex samples. Folds are constructed at the primary sampling unit (PSU) level to prevent data leakage, with Horvitz-Thompson estimators for test error. Logistic regression with group LASSO and natural splines is demonstrated using the National Survey of Family Growth (NSFG).

\item \citet{macnell_implementing_2023} compare weighting strategies for gradient boosting applied to NHANES III mortality prediction. Direct weight integration during XGBoost training is contrasted with post hoc re-evaluation using weighted outcomes.

\item \citet{wieczorek_design-based_2023} extend conformal prediction to complex surveys and provide finite sample coverage guarantees under unequal probability sampling. Applications to the Medical Expenditure Panel Survey (MEPS) and Academic Performance Index (API) show that standard conformal methods fail without design-based calibration.

\item \citet{kalpourtzi_handling_2024} compare missing data approaches in the EMENO survey (National Survey of Morbidity and Risk Factors, Greece). Multilevel multiple imputation (MI) outperforms inverse probability weighting (IPW) in bias and coverage when missingness depends on cluster-level characteristics.

\item \citet{van_alten_reweighting_2024} construct inverse probability weights (IPW) to correct volunteer bias in UK Biobank. Probit regression with least absolute shrinkage and selection operator (LASSO) estimates participation probabilities from census covariates. The resulting weights are released for UK Biobank research use.

\item \citet{wadekar_evaluating_2024} derive Horvitz-Thompson estimators for survey-weighted sensitivity and specificity. Applications to the National Survey on Drug Use and Health (NSDUH) and National Comorbidity Survey Replication (NCS-R) show that weighted metrics diverge from unweighted alternatives when weight variation is large.

\item \citet{bhaduri_review_2025} review tree-based methods for complex surveys, including classification and regression trees (CART), random forests (RF), gradient boosting (GB), and Bayesian additive regression trees (BART). Topics include weighted splitting criteria, design-based bootstrap, and cluster-aware validation.

\item \citet{falasinnu_problem_2025} develop temporal validation for chronic pain prediction using National Health Interview Survey (NHIS) data. Survey-weighted LASSO and random forest (RF) models are trained with design weights. Multiple validation scenarios assess generalizability under varying predictor availability.

\item \citet{si_machine_2025} examine whether survey weights affect prediction performance for imbalanced outcomes in NHANES. Random forest (RF), support vector machine (SVM), deep neural network (DNN), logistic regression (LR), and LASSO achieve similar individual-level accuracy regardless of weighting, raising questions about when weights are necessary for prediction.
Notably, this work concludes that complex sampling structures may be unnecessary for individual-level prediction when outcomes are highly imbalanced, because model performance is driven more by class imbalance than by survey design.

\end{itemize}


\end{document}